\documentclass[
]{ceurart}

\usepackage{listings}
\usepackage{placeins}

\usepackage{footnote}
\usepackage{hyperref}

\begin{document}

\copyrightyear{2024}
\copyrightclause{Copyright for this paper by its authors.
  Use permitted under Creative Commons License Attribution 4.0
  International (CC BY 4.0).}

\conference{3rd International Conference on Multilingual digital terminology today. Design, representation formats and management systems, June 27--28, 2024, Granada, Spain}

\title{What Generative Artificial Intelligence Means for Terminological Definitions}

\author{Antonio {San Martín}}[%
orcid=0000-0003-1732-4602,
email=antonio.san.martin.pizarro@uqtr.ca,
url=https://antoniosanmartin.info/,
]
\address{University of Quebec in Trois-Rivières, 3351, boulevard des Forges, Trois-Rivières, Quebec  G8Z 4M3, Canada}

\begin{abstract}
  This paper examines the impact of Generative Artificial Intelligence (GenAI) tools like ChatGPT on the creation and consumption of terminological definitions. From the terminologist’s point of view, the strategic use of GenAI tools can streamline the process of crafting definitions, reducing both time and effort, while potentially enhancing quality. GenAI tools enable AI-assisted terminography, notably post-editing terminography, where the machine produces a definition that the terminologist then corrects or refines. However, the potential of GenAI tools to fulfill all the terminological needs of a user, including term definitions, challenges the very existence of terminological definitions and resources as we know them. Unlike terminological definitions, GenAI tools can describe the knowledge activated by a term in a specific context. However, a main drawback of these tools is that their output can contain errors. For this reason, users requiring reliability will likely still resort to terminological resources for definitions. Nevertheless, with the inevitable integration of AI into terminology work, the distinction between human-created and AI-created content will become increasingly blurred. 
\end{abstract}



\maketitle

\section{Introduction}

Definitions are an essential element in terminological resources since they explain the conceptual content that a given term can convey. However, the creation of terminological definitions, especially the selection of the information to include in them, is very time-consuming. According to the classical approach\footnote{This approach is still defended in many terminology manuals \cite{Gouadec1990}, \cite{Suonuuti1997}, \cite{Pavel2001}, \cite{Dubuc2002}), the ISO 704:2022 standard on terminology work \cite{ISO2022}, and even manuals on terminological definitions (e.g., \cite{Vezina2009}, \cite{Fargas2009}).}, the characteristics featured in a definition (i.e., the necessary and sufficient characteristics of the defined concept) are universal and context-independent. However, they are most often impossible to determine objectively \cite{Temmerman2000}, and even when possible, this approach yields less useful definitions for non-experts.

According to the Flexible Terminological Definition Approach \cite{SanMartin2022a}, accounting for the role of context in meaning construction is necessary to overcome the drawbacks of the classical approach and create definitions that fulfill the user’s needs. From a cognitive linguistics perspective, terms, like any other lexical unit, do not possess meaning in themselves but are simply access points to large repositories of knowledge \cite{Evans2019}. It is context, understood broadly as any factor that can affect interpretation \cite{Kecskes2014}, that determines which segment of this knowledge (i.e., which meaning) is activated in each usage event.

All the knowledge that a term can invoke is its semantic potential \cite{Allwood2003}, \cite{Evans2009}. For instance, while the semantic potential of \emph{light fuel oil} includes all knowledge that it can activate in any context, the knowledge it conveys in a specific context, like in Article 20 of the Canadian Greenhouse Gas Pollution Pricing Act, is an instance of meaning. Describing a term’s semantic potential in a definition is not feasible because of its vastness. Moreover, explaining meanings is not the goal of definitions, as meanings are transient \cite{SanMartin2016}. When terminologists craft definitions, they must select the most relevant information from the term’s huge semantic potential based on contextual constraints (linguistic, thematic, cultural, ideological, geographical, and chronological \cite{SanMartin2022b}) and functional constraints (user needs, resource characteristics, etc.). Applying contextual constraints to a term's semantic potential results in a specific subset known as a premeaning, which is what a terminological definition describes.

A premeaning is a conceptualization unit halfway between a term’s semantic potential and meanings in particular usage events \cite{Croft2004}. An example of a premeaning is the knowledge activated by \emph{eutrophication} in the domain of Wastewater Treatment in a Jamaican geographical context. This contrasts with its broader semantic potential and with all its possible narrower meanings (e.g., its meaning in a tweet published by the UN Environment Programme on January 31, 2021).

If crafting traditional definitions is labor-intensive, the consideration of contextual and functional constraints makes the task even more time-consuming. This is an important barrier to the creation of flexible terminological definitions. Generative Artificial Intelligence (GenAI) tools, especially those powered by Large Language Models (LLMs) such as ChatGPT\footnote{Throughout the article, we refer to ChatGPT to represent all LLM-based tools, as it is the most popular at the time of writing. However, most observations presented here also apply to other GenAI chatbots (i.e. Google Gemini or Anthropic’s Claude). It is important to note that while terminologists currently mainly use LLMs through chatbots, GenAI integration into various applications, such as corpus analysis tools and terminology database managers, is to be expected.}, can remove these barriers by reducing the time and effort required to create them. However, the impact of GenAI can extend well beyond this, as it can profoundly transform the methods and purposes underlying the creation and consumption of terminological definitions.

With varying levels of reliability, ChatGPT can fulfill all the terminological needs of a user (interlingual equivalents, denominative variants, collocations, examples of use, pronunciation, etymology, etc.), including term definitions. ChatGPT can also assist terminologists in their work. Given this enormous potential, we can wonder, as \cite{DeSchryver2023} does for Lexicography, whether ChatGPT will replace terminological resources, terminologists, and the tools used for terminology work. Still another question is if ChatGPT will replace terminological definitions, the terminologists that write them, and the tools used for definition writing. While it is currently not possible to answer these questions because of the quickly evolving nature of GenAI, in the following sections, we explain some factors that may influence the future of terminological definitions in the face of AI.

\section{GenAI and the future of terminological definitions}

A query to ChatGPT has significant advantages over consulting a definition in a terminological resource. The main one is that while a definition can only describe a premeaning, ChatGPT can describe meaning, i.e. the knowledge activated by a term in a specific usage event. A user can ask ChatGPT to explain the meaning of a term in a specific context (Example 1)\footnote{Examples of ChatGPT prompts and responses are included in the appendix.}. ChatGPT can also tailor its responses based on the user's needs. Users can request term definitions within particular contexts --- such as a text, an image, or certain situation --- and receive answers suited to their existing knowledge (Examples 2-4).

The chat interface enables the user to ask follow-up questions to seek clarifications, request examples, etc. This interactive approach enhances the user's understanding of terms in a way that definitions cannot. Additionally, despite the limitations in how ChatGPT currently accesses online information, it can offer up-to-date information, unlike terminological definitions that may become obsolete over time. Finally, ChatGPT stands out for its user-friendliness, and it can answer questions on virtually any subject, obviating the need for users to preliminarily identify the relevant resource for their needs.

However, from the perspective of a user needing a definition, ChatGPT has certain drawbacks. More specifically, its responses may contain factual, logical, and linguistic errors, among other types \cite{Borji2023}\footnote{The problem of erroneous responses is more pronounced when interacting with ChatGPT in languages other than English \cite{OpenAI2023}. Responses in other languages may also exhibit interference from English, especially in language-related queries \cite{Jakubicek2023}.}. They can also be biased and discriminatory, and ChatGPT can even hallucinate, i.e., fabricate incorrect information. Another problem related to defining is that ChatGPT has difficulty with handling sense splitting and lumping \cite{Jakubicek2023}. This can lead to the omission of senses or the unnecessary subdivision of a single sense when prompted to create a definition. However, when provided with sufficient context, ChatGPT generally manages to select the most relevant one.

Furthermore, ChatGPT generates different responses to identical queries, which is a challenge in situations requiring a single unchanging definition. Lastly, its environmental footprint \cite{Li2023}, as well as the ethical and legal issues arising from its training on copyrighted materials \cite{Lucchi2023}, can be major drawbacks for some users.

In the face of these limitations, the main advantage of a terminological definition inserted in a resource lies in its reliability. In fact, in our view, reliability is the main reason why terminological resources, or at least some of them, will continue to be necessary despite GenAI. Nevertheless, ChatGPT’s unreliability is likely to pose a challenge only for some users in particular contexts. For instance, a translator reading a blog post for pleasure may resort to ChatGPT to understand an unfamiliar term because of its advantages and the low stakes of potential errors. However, for professional tasks, they are more likely to directly consult a terminological resource. Some will also probably turn to ChatGPT first and subsequently verify the information with reliable sources.

Some definitions have a higher chance of surviving GenAI than others. One case is the definitions with a prescriptive or standardizing orientation or those aiming to explain how terms are to be interpreted in specific contexts. These definitions can be found in standards, patents, legislation, or termbases reflecting an organization's own terminology. These definitions will not probably be replaced by AI because they intend to reflect a consensus among human beings. Another case is those definitions provided in some publications, especially pedagogical ones, to explain terms mentioned throughout the main text. They are typically found in footnotes, information boxes, or glossaries found at the end of the publication. As in-text definitions that authors provide when a term is mentioned for the first time, these definitions can survive AI because they have the advantage of being accessible to the user exactly when they are needed.

As for the potential extinction of the professional terminologists due to AI advancements, it is noteworthy that the above-mentioned definitions that are less susceptible to being replaced by AI are often not created by terminologists. However, as already discussed, terminological resources that include definitions created by terminologists are not likely to disappear. Nevertheless, their number may decrease as they become less commercially viable or lose public funding. It is believed that AI could fulfill their role. Additionally, as discussed in Section \ref{section:3}, the boundary between human-created terminological resources and definitions and those generated by AI will become increasingly blurred.

\section{AI-assisted terminography}
\label{section:3}

Given the expected pervasiveness of AI, terminologists will benefit from varying degrees of assistance from AI technologies, in what can be termed \emph{AI-assisted terminography}. This would include all forms of terminographic work where AI supports the terminologist. A particularly relevant type, especially for terminological definition writing, is post-editing terminography (an adaptation of the term \emph{post-editing lexicography} \cite{Jakubicek2018}) in which the machine produces a terminographic item (e.g., a definition) and then the terminologist corrects or refines it.

\subsection{Post-editing terminography}

In a post-editing terminography context, the strategic use of ChatGPT can enable terminologists to create definitions faster without compromising quality. This efficiency will allow terminological resources to remain up-to-date and comprehensive, and at the same time, keep pace with the swift emergence of new terms and concepts. However, this is not without its challenges.

Firstly, the potential for errors and biases requires careful verification, which can be as labor-intensive as crafting the definition from scratch. Productivity pressures could lead, in some cases, to erroneous definitions generated by ChatGPT being published without proper validation.

Furthermore, ChatGPT's responses occasionally constitute plagiarism \cite{Lee2023} or reproduce copyrighted content verbatim \cite{Mckean2023}. Unless ChatGPT bases its answer on an Internet search, it cannot typically identify its sources, and, when prompted, it may generate fictitious ones. These issues present legal and ethical concerns and complicate the task of assessing the reliability of the information provided by ChatGPT. This contrasts with corpus-based post-editing lexicography/terminography tools that enable users to consult the corresponding concordances, like the integration between Sketch Engine and Lexonomy \cite{Jakubicek2018}.

In AI-powered post-editing terminography, terminologists should be aware of the types of potential errors and remain vigilant to identify and correct them as needed. Furthermore, the post-editing effort that ChatGPT-generated definitions require can be reduced by employing optimized prompts\footnote{\cite{Lew2023} is an example of prompt optimization lexicographical definition writing.}, where terminologists specify as much as possible the characteristics of the definition.

A definition can be obtained by ChatGPT by prompting simply "Define" followed by the term to be defined. However, this approach tends to yield a lengthy encyclopedic answer (Example 5). Using “Write a terminological definition of…” normally yields a more post-editable result (Example 6). However, ChatGPT does not always follow the basic rules of definition writing, such as not repeating the defined term at the beginning of the definition. This can be solved by asking ChatGPT not to mention the defined term at the beginning of the definition (Example~7)\footnote{Asking ChatGPT to produce a lexicographic definition tends to yield a complete lexicographic entry.}.

If no contextual constraints are provided, ChatGPT typically defines terms by encompassing their most common conceptualizations across relevant knowledge domains. Furthermore, because of the nature of its training data, these definitions often tend to reflect a Western-centric perspective. Specifying in the prompt as many contextual constraints as possible as well as functional ones (target user of the definition, resource in which it will be inserted, etc.) can reduce the subsequent post-editing effort (Example 8).

When dealing with polysemic terms, it can be strategic to instruct ChatGPT to provide multiple definitions to define the different associated concepts. Without clear directives, ChatGPT might either omit certain concepts or merge them into a single definition. However, even with these instructions, terminologists should be aware that ChatGPT might omit relevant concepts or split concepts unnecessarily (Example 9).

ChatGPT is also capable of following definitional templates, which indicate the type of definitional features that the definitions of terms in the same category should include \cite{Faber2001} (Example 10). It is possible to provide ChatGPT with either empty or filled-in templates. For example, a template can indicate that in the case of antihypertensive drugs, the definitions should contain the route of administration. The terminologist can specify which are the routes of administration of each drug to ChatGPT or let ChatGPT fill this information in by itself based on its knowledge. Obviously, the more information the terminologist provides, the less risk of errors and hallucinations.

It should be noted that post-editing terminological definitions is currently only viable in high-resource languages, as the quality of ChatGPT-generated definitions in less-resourced languages may not be adequate for post-editing. Likewise, in the case of new terms or concepts not found in the ChatGPT training data, the result could also be suboptimal, even with ChatGPT's ability to base responses on web searches. Consequently, other forms of AI-assisted terminography are essential, which in turn can also be useful for validating AI-generated definitions.

\subsection{Other forms of AI-assisted terminography}

To inform definition writing, terminologists consult reference work and specialized texts. Given that terminologists must often deal with knowledge domains that are unfamiliar to them, ChatGPT can help solve all kinds of notional doubts (Examples 11-12).

Another useful method for definition writing is the analysis of existing terminological definitions of the term to be defined (when available). It follows that segmenting and contrasting definitions yield structured semantic information \cite{Faber1999}, and repeated features are likely to be relevant \cite{Faber2002}. While ChatGPT is currently not more helpful than a search engine to collect existing definitions, it is certainly helpful in analyzing them. After feeding ChatGPT a list of definitions of the same term with their source, ChatGPT can be asked to extract a list of semantic traits from them (Example 13). Once the list has been generated, the user can ask ChatGPT to perform a number of operations, such as checking the accuracy of each semantic feature (Example 14), or ranking features according to their relevance in relation to certain contextual constraints (Example 15). ChatGPT can assist in identifying gaps in the definitions by highlighting any missing information (Example 16). ChatGPT can also write a post-editable definition using the list of semantic traits and justify it based on the list (Example 17).

ChatGPT can also support definition writing by assisting with corpus analysis. Manually analyzing concordance lines to find definitional information is extremely time-consuming. ChatGPT can be fed concordance lines and asked to extract semantic information for definitional purposes. ChatGPT can help analyze concordance lines in search of semantic information related to a term to be defined (Example 18). However, due to the size limitations for ChatGPT prompts, if too many concordance lines are available for the term to be defined, the terminologist must pre-select relevant ones before submitting them to ChatGPT, akin to how they choose specific concordance lines for manual analysis in traditional corpus analysis.

An efficient way of selecting concordance lines to be analyzed is word sketches, which are automatic corpus-derived summaries of a word’s behavior created by Sketch Engine \cite{Kilgarriff2004}. They list words that appear in the corpus in a given relationship with the search word and give access to the corresponding concordance lines. Default Sketch Engine word sketches mostly represent linguistic relations (verb-object, modifiers, etc.). Although particularly useful for collocation analysis, they can also be used for semantic analysis. For instance, by choosing concordances with word sketches, we can ask ChatGPT to analyze the semantic information contained in a list of concordance lines where the target term is the object of the sentence (Example 19).

A special type of word sketches that are useful for specialized semantic analysis are semantic ones \cite{Leon-Arauz2018}, \cite{SanMartin2022c} because they extract the other terms that hold a semantic relation (hypernymy, meronymy, cause, etc.) with the search term and give access to the corresponding concordance lines. For instance, it is thus possible to ask ChatGPT to extract all the semantic information contained in a list of concordance lines where the target term holds a causal relation with other terms (Example 20).

ChatGPT can generate post-editable definitions using concordance lines as source material and justify its responses based on them (Example 21). However, manually copying and pasting concordances from corpus tools into ChatGPT can be time-consuming and inefficient. The anticipated integration of AI into corpus tools is expected to greatly streamline corpus analysis.

Finally, ChatGPT can also be a valuable tool for refining terminological definitions. After drafting a definition, terminologists can ask ChatGPT to assess and suggest enhancements. This addresses both content-related issues, such as errors and biases, and formal aspects such as spelling and stylistic choices (Example 22). Moreover, ChatGPT's ability to guess the term from a definition offers an innovative approach to evaluating the adequacy of a definition. Should ChatGPT fail to correctly identify the defined term, this may indicate that the definition is in need of further refinement, a process in which ChatGPT can also provide assistance (Example~23).

\section{Conclusions}

AI has the potential to revolutionize many aspects of human life, and terminology work is no exception. Up until now, terminological definitions have been the terminographic element that has most escaped automation \cite{Rundell2023}, but GenIA-based tools have radically changed the situation. They are now able to provide terminological definitions that many users might consider good enough, which means that they would not consult terminological resources. This shift challenges the very existence of terminological resources as we know them.

While users requiring reliability are likely to still consult terminological resources, the distinction between human-created and AI-created content will become increasingly blurred. ChatGPT and the expected emergence of AI tools tailored specifically for terminology work are set to substantially boost the efficiency of terminologists who must write definitions. This increase in productivity will not only allow for the prompt update of terminological resources to reflect knowledge advancement but can also lead to an improvement in quality by facilitating the creation of flexible definitions. This means that crafting definitions that accurately reflect the contextual variability of terms will no longer be constrained by the lengthy and labor-intensive nature of the process.

Considering these advantages, the integration of AI into terminology work appears inevitable. As AI technologies grow increasingly sophisticated, the gap in quality between AI-generated and human-created content is likely to diminish. To remain competitive and relevant against AI tools such as ChatGPT, terminologists will need to embrace AI and develop the necessary skills for its efficient integration into their workflow. Just as using computers for terminography has become a standard practice, we foresee a future where AI-assisted terminography will be synonymous with terminography itself.


\begin{acknowledgments}
This research was carried out as part of projects 2020-NP-267503 funded by Quebec’s Society and Culture Research Fund, 430-2023-0248 funded by the Social Sciences and Humanities Research Council of Canada, and PID2020-118369GBI00 funded by the Spanish Ministry of Science and Innovation.
\end{acknowledgments}

\bibliography{bibliomdtt}

\appendix
\section{Appendix}

All the answers to the prompts below were obtained with ChatGPT 4 in March 2024.

\subsection*{Example 1: Asking ChatGPT to provide the definition of a term in a specific textual context}
\label{example:1}

\noindent \framebox{\textbf{User}}\\
\noindent What does “work” mean in “For a particle acted upon only by conservative forces, the work done is path-independent.”?\\

\noindent \framebox{\textbf{ChatGPT}}\\
\noindent In the context of physics, "work" refers to the transfer of energy to an object via a force causing the object to move. The work done on the object is calculated as the product of the force applied to the object, the distance over which the force is applied, and the cosine of the angle between the force and the direction of movement.\\

\noindent When it is said that "for a particle acted upon only by conservative forces, the work done is path-independent," it means that the total work done on the particle as it moves from one point to another is the same, regardless of the path taken between these points. Conservative forces, such as gravitational and electrostatic forces, have this property because the work done by these forces depends only on the initial and final positions of the particle, not on the specific path taken between these positions. Therefore, the total work done by conservative forces on a particle in moving from one point to another is determined solely by these points and is independent of the trajectory followed by the particle.

\subsection*{Example 2: Asking ChatGPT to provide the definition of a term in a specific textual context including background information about the user and the text}
\label{example:2}

\noindent \framebox{\textbf{User}}\\ 
\noindent I am a terminologist, and I am reading an introductory book on statistics for corpus analysis. Can you explain what “predictive modeling” means in this sentence of the book: “Several statistical methods can be used for predictive modeling in corpus linguistics”?\\

\noindent \framebox{\textbf{ChatGPT}}\\
\noindent Predictive modeling in the context of corpus linguistics refers to the use of statistical methods to analyze textual data (corpus) in order to make predictions about future instances of language use or to generalize about language patterns. This involves building a model that can predict an outcome based on the features extracted from the text. Here’s how it typically works:

    \begin{enumerate}
    \item \textbf{Data Collection and Preparation}: A large corpus of text is gathered and prepared for analysis. This could include cleaning the data, annotating it, and structuring it in a way that can be used for statistical analysis.
    \item \textbf{Feature Extraction}: Characteristics (features) of the text are identified and extracted. These features might include word frequency, grammatical structures, n-grams (sequences of `n` words), syntactic patterns, etc.
    \item \textbf{Model Training}: The extracted features are used to train a statistical model. This involves using a portion of the corpus to teach the model what the outcomes should be based on the features. Common methods include regression analysis, decision trees, machine learning algorithms, and more.
    \item \textbf{Testing and Validation}: The model is then tested with another portion of the corpus (not used in the training phase) to check its accuracy and ability to predict outcomes. This step is crucial to validate that the model can generalize beyond the specific data it was trained on.
    \item \textbf{Prediction}: Once the model is validated, it can be used to make predictions. For example, it could predict the likelihood of certain language structures or words appearing in similar texts, or it might be used to infer characteristics of the text's authors or the contexts in which the text is likely to occur.
    \end{enumerate}

\noindent In corpus linguistics, predictive modeling can be used for various purposes, such as identifying future language trends, automating text classification, understanding language evolution, or improving natural language processing systems.

\subsection*{Example 3: Asking ChatGPT to provide the definition of a term in an image including background information about the user\footnote{Image by Chrislb (Diagram of an artificial neuron). Licensed under the Creative Commons Attribution-Share Alike 3.0 Unported license. URL: https://commons.wikimedia.org/wiki/File:ArtificialNeuronModel\_english.png}}
\label{example:3}

\noindent \framebox{\textbf{User}}\\
\noindent What does “weight” mean in this image? Explain in simple terms because I have no background in artificial intelligence.\\\\
 \framebox{\includegraphics[width=\linewidth]{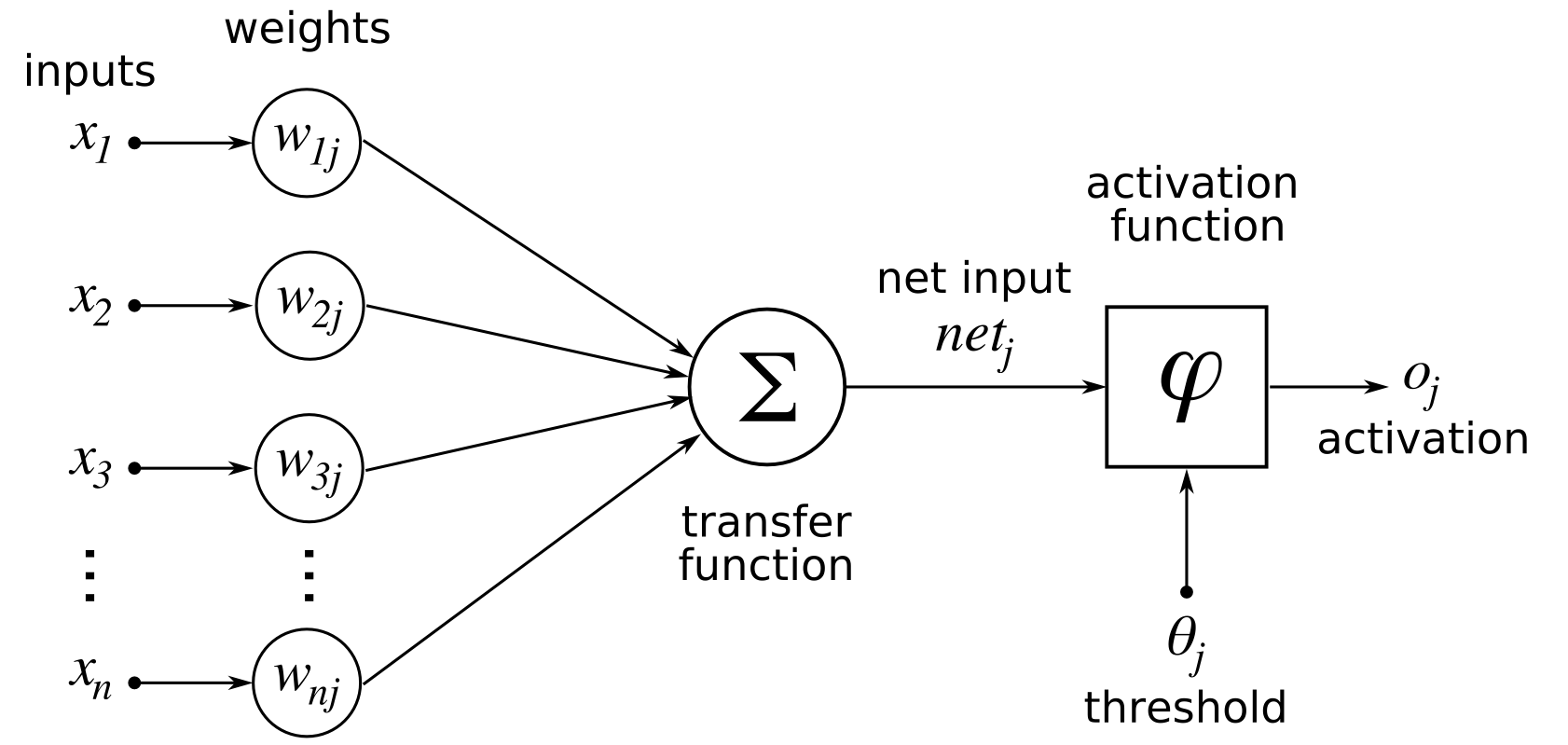}}\\

\noindent \framebox{\textbf{ChatGPT}}\\ 
\noindent The image you've uploaded shows a diagram of a single neuron within a neural network, which is a fundamental concept in artificial intelligence. Here's a simple breakdown of what "weight" means in this context:\\ 

\noindent Think of a neural network like a team of decision-makers, where each "neuron" is an individual decision-maker. Each decision-maker (neuron) gets a set of inputs, which are pieces of information. The "weights" are like the importance each decision-maker assigns to each piece of information when making a decision.\\ 

\noindent For example, if the decision is about whether to take an umbrella when leaving the house, one piece of input information might be the weather forecast saying it's going to rain. If a particular neuron thinks the weather forecast is very important, it will assign a high weight to that information. On the other hand, if another input is the fact that it rained yesterday, and the neuron doesn't think that's very important, it might assign a low weight to that information.\\ 

\noindent So, "weights" adjust how much influence each piece of input information has on the decision each neuron makes. After considering all the inputs and their respective weights, the neuron then decides what output to provide, which contributes to the final decision made by the whole neural network.

\subsection*{Example 4: Asking ChatGPT to provide the definition of a term in a specific situation}
\label{example:4}

\noindent \framebox{\textbf{User}}\\
\noindent My cousin is a truck driver and said something about a “sideboard”. What does it mean?\\

\noindent \framebox{\textbf{ChatGPT}}\\
\noindent A "sideboard" in the context of truck driving usually refers to an attachment on the sides of a truck bed or trailer that helps contain the load and prevent it from spilling or shifting during transport. These sideboards are often made of wood or metal and can be either fixed or removable. They effectively increase the height of the sides of the truck bed or trailer, allowing for more secure transportation of larger or more voluminous cargo.

\subsection*{Example 5: Asking ChatGPT to provide the definition of a term}
\label{example:5}

\noindent \framebox{\textbf{User}}\\
\noindent Define “spectroscopy”.\\

\noindent \framebox{\textbf{ChatGPT}}\\
\noindent Spectroscopy is a scientific technique used to analyze the properties of light and other forms of electromagnetic radiation as they interact with matter. It involves measuring the spectrum of light—that is, the distribution of light intensity with respect to wavelength or frequency—from an object or sample. By examining these spectra, scientists can infer various characteristics of the matter being studied, such as its composition, temperature, density, and chemical structure.\\

\noindent Spectroscopy is used in many fields, including chemistry, physics, astronomy, and biology, to identify substances, study atomic and molecular structures, and understand physical processes. Different types of spectroscopy, such as infrared, ultraviolet, and X-ray spectroscopy, utilize different regions of the electromagnetic spectrum to provide information about the matter under investigation.

\subsection*{Example 6: Asking ChatGPT to provide the terminological definition of a term}
\label{example:6}

\noindent \framebox{\textbf{User}}\\
\noindent Write a terminological definition of “spectroscopy”.\\

\noindent \framebox{\textbf{ChatGPT}}\\
\noindent Spectroscopy is a scientific technique used to analyze the interactions between matter and electromagnetic radiation. It involves measuring the absorption, emission, or scattering of light by materials to determine their composition, structure, and physical properties. Spectroscopy can identify chemical elements and compounds, determine concentrations, and reveal molecular energy levels and mechanisms. This method is fundamental in various fields, including chemistry, physics, astronomy, and environmental science, offering insights into the characteristics of different substances and celestial objects.

\subsection*{Example 7: Asking ChatGPT to provide the terminological definition of a term specifying that the defined term should not be mentioned at the beginning of the definition}
\label{example:7}

\noindent \framebox{\textbf{User}}\\
\noindent Write a terminological definition of "spectroscopy". Do not mention the defined term at the beginning of the definition.\\

\noindent \framebox{\textbf{ChatGPT}}\\
\noindent A scientific technique that involves analyzing the interaction between matter and electromagnetic radiation to identify, quantify, and study the properties of substances. This method can be applied across a broad spectrum of wavelengths, including infrared, visible, ultraviolet, and X-ray regions. It is used to infer the physical properties, chemical composition, and structure of the analyzed material by examining its emitted, absorbed, or scattered light.

\subsection*{Example 8: Asking ChatGPT to provide the terminological definition of a term specifying contextual and functional constraints}
\label{example:8}

\noindent \framebox{\textbf{User}}\\
\noindent Write a terminological definition of “hydraulic fracturing” intended for high school students from the thematic perspective of Energy Engineering and the geographic perspective of Canada. Do not mention the defined term at the beginning of the definition. The definition will be inserted in a specialized dictionary for high school students.\\

\noindent \framebox{\textbf{ChatGPT}}\\
\noindent A technique used in the energy industry to extract oil and natural gas from deep underground, this process involves injecting high-pressure fluid into rock formations. This fluid creates fractures in the rock, allowing the trapped oil or gas to flow to the surface. In Canada, particularly in regions like Alberta and British Columbia, it has significantly contributed to the country's energy production, tapping into vast reserves that were previously inaccessible. This method has transformed the energy landscape, making Canada one of the top producers of natural gas and oil in the world.

\subsection*{Example 9: Asking ChatGPT to provide the terminological definition of a term specifying contextual constraints and allowing for more than one definition if the term is polysemic}
\label{example:9}

\noindent \framebox{\textbf{User}}\\
\noindent Write a terminological definition of “feed” from the thematic perspective of Agriculture. If the term is polysemic, write one definition per concept associated to the term.\\

\noindent \framebox{\textbf{ChatGPT}}\\
\noindent In agriculture, \textit{feed} refers to the food given to domestic animals, particularly livestock, in the course of animal husbandry. The term is polysemic and has several definitions based on its different uses within this thematic perspective:

    \begin{enumerate}
    \item \textbf{Animal Nutrition:} As a noun, \textit{feed} encompasses all types of food given to animals, including hay, grain, and commercially prepared animal feed. This type of feed is specifically formulated to provide essential nutrients that ensure the health, growth, and productivity of livestock and other animals.
    \item \textbf{Feeding Process:} As a verb, \textit{feed} means the act of providing food to animals. This involves the physical process of delivering food to animals, which can vary from manual distribution to automated feeding systems in a controlled and often scheduled manner to meet the nutritional needs of animals.
    \item \textbf{Economic Good:} In an economic context within agriculture, \textit{feed} refers to the commodity that is bought and sold for the purpose of animal consumption. The quality, composition, and price of feed can significantly impact the agricultural sector, especially in livestock production.
    \end{enumerate}

\subsection*{Example 10: Asking ChatGPT to provide the definition of several terms based on a definitional template}
\label{example:10}

\noindent \framebox{\textbf{User}}\\
\noindent I have the following template to create the definition of concepts belonging to the category HARD COASTAL DEFENSE STRUCTURE:\\

\noindent type\_of:

\noindent located\_at:

\noindent made\_of:

\noindent has\_function:\\

\noindent This is the template as applied to the concept GROYNE:\\

\noindent Concept: GROYNE\\
 
\noindent Definition: hard coastal defense structure made of concrete, wood, steel and/or rock perpendicular to the shoreline built to protect a shore area, retard littoral drift, reduce longshore transport, and prevent beach erosion.\\

\noindent Template:

\noindent type\_of: hard coastal defense structure

\noindent located\_at: perpendicular to shoreline

\noindent made\_of: concrete, wood, steel, and/or rock

\noindent has\_function: shore protection, littoral drift retardation, longshore transport reduction, and beach erosion prevention\\

\noindent Complete the template of the concepts JETTY, MOLE, and DIKE one by one. And write the definition of each concept after its template.\\

\noindent \framebox{\textbf{ChatGPT}}\\
\noindent Let's start with the concept of a JETTY.\\

\noindent\textbf{Concept: JETTY}

\noindent Template:
\begin{itemize}

\item type\_of: hard coastal defense structure 

\item located\_at: alongside shorelines or entrances to harbors and ports 

\item made\_of: stone, concrete, and/or wood 

\item as\_function: protection of harbors and coastline, control of sediment transport, and provision of safe berthing areas
\end{itemize}

\noindent Definition: A jetty is a hard coastal defense structure constructed of stone, concrete, or wood, extending from the shore into the water, often at harbor entrances or river mouths. It is designed to protect the shoreline and harbor areas by controlling sediment transport, preventing shoreline erosion, and providing safe mooring areas. 

\noindent Next, we'll cover the concept of a MOLE.\\

\noindent \textbf{Concept: MOLE} 

\noindent Template: 
\begin{itemize}

\item type\_of: hard coastal defense structure 

\item located\_at: extending into the sea from a coast or harbor 

\item made\_of: large stones, concrete, and/or wood 

\item has\_function: harbor protection, wave break, and creation of calm waters for docks and marinas

\end{itemize}

\noindent Definition: A mole is a massive hard coastal defense structure made of large stones, concrete, or wood, extending from the coast or harbor into the sea. Its primary functions are to protect harbors, act as a breakwater to reduce the force of waves, and create calm waters for docks, marinas, and anchorages. 

\noindent Finally, let's define a DIKE.\\

\noindent \textbf{Concept: DIKE}

\noindent Template: 
\begin{itemize}

\item type\_of: hard coastal defense structure 

\item located\_at: along coastlines or riverbanks 

\item made\_of: earth, clay, concrete, and/or rock 

\item has\_function: flood prevention, erosion control, and land reclamation
\end{itemize}

\noindent Definition: A dike is a hard coastal defense structure built along coastlines or riverbanks, constructed from materials such as earth, clay, concrete, or rock. It is designed to prevent flooding, control erosion, and reclaim land from bodies of water, thereby protecting and maintaining the integrity of inland areas.

\subsection*{Example 11: Asking ChatGPT to explain the difference between two terms}
\label{example:11}

\noindent \framebox{\textbf{User}}\\
\noindent What is the difference between “quantum decoherence” and “wave function collapse”?\\

\noindent \framebox{\textbf{ChatGPT}}\\
\noindent Quantum decoherence and wave function collapse are related concepts in quantum mechanics, but they refer to different processes:

\begin{itemize} 

  \item \textbf{Quantum Decoherence}: This is the process by which a quantum system loses its quantum properties and behavior, transitioning towards classical behavior due to the interaction with its environment. Decoherence explains how the superposition of states (where a system exists in multiple states simultaneously) appears to reduce to a single outcome. However, decoherence does not predict which outcome is realized; it only describes how coherent superpositions are lost, leaving a mixture of states rather than a single definite state. Essentially, decoherence provides a mechanism for the classical-like behavior of macroscopic systems by showing how they lose their quantum coherence through interactions with their surroundings, leading to the appearance of wave function collapse without actually collapsing the wave function.

  \item \textbf{Wave Function Collapse}: This concept is used to describe the change of a quantum system from a superposition of states to a single eigenstate due to a measurement or observation. Wave function collapse is a fundamental concept in the Copenhagen interpretation of quantum mechanics, where the act of measurement forces a quantum system to settle into one of the possible eigenstates. The wave function collapse is instantaneous and discontinuous, and it selects one outcome from the probabilities described by the wave function. 

\end{itemize} 

\noindent In summary, quantum decoherence describes how interactions with the environment cause a quantum system to appear classical, leading to a loss of quantum superposition, but without specifying which state the system will end up in. On the other hand, wave function collapse refers to the process where a quantum system suddenly transitions from a superposition of states to a single definite state upon measurement. While decoherence can be seen as a part of the process leading to collapse by effectively "hiding" the other outcomes from the superposition, it does not by itself explain why a specific outcome is observed in a measurement.

\subsection*{Example 12: Asking for the explanation of a sentence}
\label{example:12}

\noindent \framebox{\textbf{User}}\\
\noindent Explain this sentence in simple terms: “Energy levels in quantum systems are generally the real eigenvalues of their Hermitian Hamiltonian”.\\

\noindent \framebox{\textbf{ChatGPT}}\\
\noindent This sentence is about how energy is described in quantum mechanics, which is the branch of physics dealing with very small particles like atoms and electrons. 

\begin{itemize} 

    \item \textbf{Quantum systems}: These are systems studied in quantum mechanics, like atoms or particles, where the rules of classical physics don't apply. 

    \item \textbf{Energy levels}: In quantum systems, energy doesn't vary smoothly but comes in discrete steps or "levels." Just like you can only climb stairs one step at a time, particles in quantum systems can only have specific energy levels. 

    \item \textbf{Real eigenvalues}: This term comes from mathematics, particularly from the study of matrices. Eigenvalues are numbers that give important information about a system. In this context, "real" means these numbers are not complex (i.e., they don't involve imaginary numbers, which are numbers involving the square root of negative one). 

    \item \textbf{Hermitian Hamiltonian}: This is a fancy way of saying a specific type of mathematical function or operator that is used in quantum mechanics to describe the total energy of a system. "Hermitian" means it has certain symmetry properties, and "Hamiltonian" refers to the total energy of the system. 

\end{itemize} 

\noindent So, putting it all together, the sentence means that in quantum mechanics, the specific energy values that a quantum system can have are determined by solving a mathematical problem involving the system's Hamiltonian, which is a function that describes its total energy. The solutions to this problem are the energy levels, and they are real numbers, not complex numbers.

\subsection*{Example 13: Asking ChatGPT to provide a list of all the semantic features in a set of definitions of the same term}
\label{example:13}

\noindent \framebox{\textbf{User}}\\ 
\noindent Make a list of all the semantic features of "radiative forcing" contained in these definitions. Each semantic feature should be a sentence. Indicate in brackets the source that contains each semantic trait. Be exhaustive: include all possible traits, even those mentioned in only one definition, but if a trait is mentioned in more than one definition, list all sources.\\

\noindent A change in average net radiation at the top of the troposphere (known as the tropopause) because of a change in either incoming solar or exiting infrared radiation. A positive radiative forcing tends on average to warm the earth's surface; a negative radiative forcing on average tends to cool the earth's surface. Greenhouse gases, when emitted into the atmosphere, trap infrared energy radiated from the earth's surface and therefore tend to produce positive radiative forcing. (EIA: \url{https://www.eia.gov/tools/glossary/})\\

\noindent A change in the balance between incoming solar radiation and outgoing infrared (i.e., thermal) radiation.  Without any radiative forcing, solar radiation coming to the Earth would continue to be approximately equal to the infrared radiation emitted from the Earth.  The addition of greenhouse gases to the atmosphere traps an increased fraction of the infrared radiation, reradiating it back toward the surface of the Earth and thereby creates a warming influence. (UNFCCC: \url{https://unfccc.int/resource/cd_roms/na1/ghg_inventories/english/8_glossary/Glossary.htm})\\

\noindent Change in the balance between incoming solar radiation and outgoing infrared radiation. Causes include internal changes and external forcing, such as changes in solar output or carbon dioxide concentrations. Without any radiative forcing, solar radiation coming to the Earth would approximately equal to the infrared radiation emitted from Earth. A positive forcing warms the Earth, a negative forcing cools it. (CARA: \url{https://web.archive.org/web/20060901114821/http://www.cara.psu.edu/tools/glossary.asp})\\

\noindent A change in the balance between incoming solar radiation and outgoing infra-red radiation. Without any radiative forcing, solar radiation coming to the Earth would continue to be approximately equal to the infra-red radiation emitted from the Earth. The addition of greenhouse gases traps and increased fraction of the infra-red radiation, reradiating it back toward the surface and creating a warming influence (i.e., positive radiative forcing because incoming solar radiation will exceed outgoing infra-red radiation). (EO-NASA: \url{https://earthobservatory.nasa.gov/glossary/q/s})\\

\noindent A change in average net radiation (in W m-2) at the top of the troposphere resulting from a change in either solar or infrared radiation due to a change in atmospheric greenhouse gases concentrations; perturbance in the balance between incoming solar radiation and outgoing infrared radiation. (IPCC: \url{https://archive.ipcc.ch/ipccreports/sres/aviation/index.php?idp=158})\\

\noindent The term radiative forcing refers to changes in the energy balance of the earth-atmosphere system in response to a change in factors such as greenhouse gases, land-use change, or solar radiation. The climate system inherently attempts to balance incoming (e.g., light) and outgoing (e.g. heat) radiation. Positive radiative forcings increase the temperature of the lower atmosphere, which in turn increases temperatures at the Earth's surface. Negative radiative forcings cool the lower atmosphere. Radiative forcing is most commonly measured in units of watts per square meter (W/m2). (Climate CoLab: \url{https://www.climatecolab.org/wiki/Glossary})\\

\noindent A change imposed upon the climate system which modifies the radiative balance of that system. The causes of such a change may include changes in the sun, clouds, ice, greenhouse gases, volcanic activity, and other agents. Radiative forcing is often specified as the net change in energy flux at the troposphere (watts per square meter). Radiative forcing may sometimes be referred to as external forcing or perturbations of the climate. (GCMD-NASA: \url{https://gcmd.earthdata.nasa.gov/KeywordViewer/})\\

\noindent The net change in the energy balance of the Earth system due to an external perturbation, measured in units of watts per square metre (W/m2). (CCCR: \url{https://natural-resources.canada.ca/sites/www.nrcan.gc.ca/files/energy/Climate-change/pdf/CCCR_Definitions-EN-111919.pdf})\\

\noindent Difference between incoming solar radiation on the Earth and outgoing thermal radiation from the Earth. (ISO 22948:2020: \url{https://www.iso.org/obp/ui#iso:std:iso:22948:ed-1:v1:en:term:3.5.4})\\

\noindent A measure of the influence of a particular factor (e.g. greenhouse gas (GHG), aerosol, or land use change) on the net change in the Earths energy balance. (EPA: \url{https://19january2017snapshot.epa.gov/climatechange/glossary-climate-change-terms_.html})\\

\noindent A disturbance to the earth's energy balance. (Termium Plus: \url{https://www.btb.termiumplus.gc.ca/})\\

\noindent A perturbation of the global radiation balance. This may be human-induced or natural. (LennTech: \url{https://www.lenntech.com/greenhouse-effect/climate-change-glossary.htm})\\

\noindent A change in the balance between incoming solar radiation and outgoing infrared radiation. Without any radiative forcing, solar radiation coming to the Earth would continue to be approximately equal to the infrared radiation emitted from the Earth. The addition of greenhouse gases traps an increased fraction of the infrared radiation, radiating it back toward the surface and creating a warming influence (i.e., positive radiative forcing because incoming solar radiation will exceed outgoing infrared radiation). (NAL: \url{https://agclass.nal.usda.gov/vocabularies/nalt/concept?uri=https://lod.nal.usda.gov/nalt/281696})\\

\noindent Changes in energy fluxes F (in W/m2) at the top-of-the-atmosphere that cause climate change. It is defined as positive (negative) if it leads to warming (cooling). The radiative forcing for a doubling of CO2 is $\Delta$F2xCO2 = 3.7 W/m2. Other examples are increased solar radiation (positive), increased aerosols (negative) or increased surface albedo (negative), e.g. due to land use changes. (OSU: \url{https://open.oregonstate.education/climatechange/back-matter/glossary/})\\

\noindent The effect of greenhouse gases on the radiation of heat in the atmosphere. (Wiktionary: \url{https://en.wiktionary.org/})\\

\noindent Radiative forcing is the change in energy flux in the atmosphere caused by natural or anthropogenic factors of climate change as measured in watts per meter squared. (Wikipedia: \url{https://en.wikipedia.org/})\\

\noindent The difference between sunlight absorbed by Earth and energy radiated back to space, disrupted by greenhouse gas levels. (Footprint Consulting: \url{https://footprintconsulting.org/climate-knowledge-base/})
 \\

\noindent \framebox{\textbf{ChatGPT}}

\begin{itemize}
\item Radiative forcing occurs at the top of the troposphere, specifically at the tropopause. (EIA, IPCC) 

\item It involves a change in average net radiation due to variations in incoming solar or outgoing infrared radiation. (EIA, IPCC) 

\item Positive radiative forcing tends to warm the Earth's surface, while negative radiative forcing tends to cool it. (EIA, CARA, OSU) 

\item Greenhouse gases trap infrared energy radiated from the Earth's surface, leading to positive radiative forcing. (EIA, UNFCCC, EO-NASA) 

\item Radiative forcing affects the balance between incoming solar radiation and outgoing infrared radiation. (UNFCCC, CARA, EO-NASA, IPCC, NAL) 

\item Without radiative forcing, solar radiation coming to Earth would be approximately equal to the infrared radiation emitted from Earth. (UNFCCC, CARA, EO-NASA, NAL) 

\item Causes of radiative forcing include internal changes and external factors like solar output or carbon dioxide concentrations. (CARA) 

\item Greenhouse gases increase the fraction of trapped infrared radiation, reradiating it back toward the Earth's surface, creating a warming effect. (UNFCCC, EO-NASA, NAL) 

\item It is measured in watts per square meter (W/m\^2). (IPCC, Climate CoLab, GCMD-NASA, CCCR, OSU) 

\item Radiative forcing reflects changes in the energy balance of the earth-atmosphere system due to various factors such as greenhouse gases and solar radiation. (Climate CoLab, GCMD-NASA) 

\item The climate system inherently tries to balance incoming and outgoing radiation. (Climate CoLab) 

\item It can increase or decrease the temperature of the lower atmosphere, affecting the Earth's surface temperature. (Climate CoLab) 

\item Radiative forcing can be imposed by changes in the sun, clouds, ice, greenhouse gases, volcanic activity, and other agents. (GCMD-NASA) 

\item It is often specified as the net change in energy flux at the troposphere. (GCMD-NASA) 

\item It may sometimes be referred to as external forcing or perturbations of the climate. (GCMD-NASA) 

\item Represents the net change in the Earth system's energy balance due to external perturbation. (CCCR) 

\item Described as the difference between incoming solar radiation on the Earth and outgoing thermal radiation from the Earth. (ISO 22948:2020) 

\item Indicates the influence of a particular factor on the net change in Earth's energy balance. (EPA) 

\item Considered a disturbance to the Earth's energy balance. (Termium Plus) 

\item Perturbs the global radiation balance, either human-induced or natural. (LennTech) 

\item Changes in energy fluxes at the top-of-the-atmosphere cause climate change, defined as positive for warming and negative for cooling. (OSU) 

\item Relates to the effect of greenhouse gases on the radiation of heat in the atmosphere. (Wiktionary) 

\item It is the change in energy flux in the atmosphere caused by natural or anthropogenic factors. (Wikipedia) 

\item Involves the disruption of the balance between absorbed sunlight by Earth and energy radiated back to space due to greenhouse gas levels. (Footprint Consulting)
\end{itemize}

\subsection*{Example 14: Continuation of the conversation in Example 13. Asking ChatGPT to verify the accuracy of each semantic trait}
\label{example:14}

\noindent \framebox{\textbf{User}}\\
\noindent Verify exhaustively the accuracy of each semantic trait. Copy each semantic trait, add the source(s), and then label it as accurate or inaccurate according to your knowledge. If the semantic trait is accurate, do not write anything else. If the semantic trait is inaccurate, write an explanation and propose a corrected version of the trait.\\

\noindent \framebox{\textbf{ChatGPT}}\\ 

\begin{enumerate}

\item \textbf{Radiative forcing occurs at the top of the troposphere, specifically at the tropopause. (EIA, IPCC)}

   Accurate

\item \textbf{It involves a change in average net radiation due to variations in incoming solar or outgoing infrared radiation. (EIA, IPCC)} 

   Accurate

\item \textbf{Positive radiative forcing tends to warm the Earth's surface, while negative radiative forcing tends to cool it. (EIA, CARA, OSU)}   

   Accurate

\item \textbf{Greenhouse gases trap infrared energy radiated from the Earth's surface, leading to positive radiative forcing. (EIA, UNFCCC, EO-NASA)}   

   Accurate

\item \textbf{Radiative forcing affects the balance between incoming solar radiation and outgoing infrared radiation. (UNFCCC, CARA, EO-NASA, IPCC, NAL)}   

   Accurate

\item \textbf{Without radiative forcing, solar radiation coming to Earth would be approximately equal to the infrared radiation emitted from Earth. (UNFCCC, CARA, EO-NASA, NAL)}   

   Accurate

\item \textbf{Causes of radiative forcing include internal changes and external factors like solar output or carbon dioxide concentrations. (CARA)}   

   Accurate

\item \textbf{Greenhouse gases increase the fraction of trapped infrared radiation, reradiating it back toward the Earth's surface, creating a warming effect. (UNFCCC, EO-NASA, NAL)}   

   Accurate

\item \textbf{It is measured in watts per square meter (W/m\^2). (IPCC, Climate CoLab, GCMD-NASA, CCCR, OSU)}   

   Accurate

\item \textbf{Radiative forcing reflects changes in the energy balance of the earth-atmosphere system due to various factors such as greenhouse gases and solar radiation. (Climate CoLab, GCMD-NASA)}   

    Accurate

\item \textbf{The climate system inherently tries to balance incoming and outgoing radiation. (Climate CoLab)}   

    Accurate

\item \textbf{It can increase or decrease the temperature of the lower atmosphere, affecting the Earth's surface temperature. (Climate CoLab)}   

    Accurate

\item \textbf{Radiative forcing can be imposed by changes in the sun, clouds, ice, greenhouse gases, volcanic activity, and other agents. (GCMD-NASA)}   

    Accurate

\item \textbf{It is often specified as the net change in energy flux at the troposphere. (GCMD-NASA)}   

    This trait is slightly inaccurate because radiative forcing is commonly measured at the top of the atmosphere, not just at the troposphere.   

    Corrected version: Radiative forcing is often specified as the net change in energy flux at the top of the atmosphere.

\item \textbf{It may sometimes be referred to as external forcing or perturbations of the climate. (GCMD-NASA)}   

    Accurate

\item \textbf{Represents the net change in the Earth system's energy balance due to external perturbation. (CCCR)}   

    Accurate

\item \textbf{Described as the difference between incoming solar radiation on the Earth and outgoing thermal radiation from the Earth. (ISO 22948:2020)}   

    Accurate

\item \textbf{Indicates the influence of a particular factor on the net change in Earth's energy balance. (EPA)}   

    Accurate

\item \textbf{Considered a disturbance to the Earth's energy balance. (Termium Plus)}   

    Accurate

\item \textbf{Perturbs the global radiation balance, either human-induced or natural. (LennTech)}   

    Accurate

\item \textbf{Changes in energy fluxes at the top-of-the-atmosphere cause climate change, defined as positive for warming and negative for cooling. (OSU)}   

    Accurate

\item \textbf{Relates to the effect of greenhouse gases on the radiation of heat in the atmosphere. (Wiktionary)}   

    Accurate

\item \textbf{It is the change in energy flux in the atmosphere caused by natural or anthropogenic factors. (Wikipedia)}   

    This trait is somewhat imprecise, as it doesn't clearly state that the change occurs at the top of the atmosphere.   

    Corrected version: Radiative forcing is the change in energy flux at the top of the atmosphere caused by natural or anthropogenic factors.

\item \textbf{Involves the disruption of the balance between absorbed sunlight by Earth and energy radiated back to space due to greenhouse gas levels. (Footprint Consulting)}   

    Accurate

\end{enumerate}

\subsection*{Example 15: Continuation of the conversation in Example 14. Asking ChatGPT to classify the semantic traits based on relevance}
\label{example:15}

\noindent \framebox{\textbf{User}}\\
\noindent Classify all of the traits in three categories (highly relevant, relevant, less relevant) according to the relevance to define "radiative forcing" in a terminological resource about climate change. If a trait has been corrected, use the corrected version.\\

\noindent \framebox{\textbf{ChatGPT}}\\
\noindent \textbf{Highly Relevant:}

\begin{itemize}

\item Radiative forcing occurs at the top of the troposphere, specifically at the tropopause. (EIA, IPCC) 

\item It involves a change in average net radiation due to variations in incoming solar or outgoing infrared radiation. (EIA, IPCC) 

\item Positive radiative forcing tends to warm the Earth's surface, while negative radiative forcing tends to cool it. (EIA, CARA, OSU) 

\item Radiative forcing affects the balance between incoming solar radiation and outgoing infrared radiation. (UNFCCC, CARA, EO-NASA, IPCC, NAL) 

\item Greenhouse gases trap infrared energy radiated from the Earth's surface, leading to positive radiative forcing. (EIA, UNFCCC, EO-NASA) 

\item Without radiative forcing, solar radiation coming to Earth would be approximately equal to the infrared radiation emitted from Earth. (UNFCCC, CARA, EO-NASA, NAL) 

\item It is measured in watts per square meter (W/m\^2). (IPCC, Climate CoLab, GCMD-NASA, CCCR, OSU) 

\item Represents the net change in the Earth system's energy balance due to external perturbation. (CCCR) 
\end{itemize}

\noindent \textbf{Relevant:} 
\begin{itemize}
\item Causes of radiative forcing include internal changes and external factors like solar output or carbon dioxide concentrations. (CARA) 

\item Greenhouse gases increase the fraction of trapped infrared radiation, reradiating it back toward the Earth's surface, creating a warming effect. (UNFCCC, EO-NASA, NAL) 

\item The climate system inherently tries to balance incoming and outgoing radiation. (Climate CoLab) 

\item It can increase or decrease the temperature of the lower atmosphere, affecting the Earth's surface temperature. (Climate CoLab) 

\item Radiative forcing can be imposed by changes in the sun, clouds, ice, greenhouse gases, volcanic activity, and other agents. (GCMD-NASA) 

\item It may sometimes be referred to as external forcing or perturbations of the climate. (GCMD-NASA) 

\item It is often specified as the net change in energy flux at the troposphere. (GCMD-NASA) 
\end{itemize}
 
\newpage
\noindent \textbf{Less Relevant:}
\begin{itemize}
\item Indicates the influence of a particular factor on the net change in Earth's energy balance. (EPA) 

\item Considered a disturbance to the Earth's energy balance. (Termium Plus) 

\item Perturbs the global radiation balance, either human-induced or natural. (LennTech) 

\item Changes in energy fluxes at the top-of-the-atmosphere cause climate change, defined as positive for warming and negative for cooling. (OSU) 

\item Relates to the effect of greenhouse gases on the radiation of heat in the atmosphere. (Wiktionary) 

\item It is the change in energy flux in the atmosphere caused by natural or anthropogenic factors. (Wikipedia) 

\item Involves the disruption of the balance between absorbed sunlight by Earth and energy radiated back to space due to greenhouse gas levels. (Footprint Consulting) 

\item Described as the difference between incoming solar radiation on the Earth and outgoing thermal radiation from the Earth. (ISO 22948:2020)
\end{itemize}

\subsection*{Example 16: Continuation of the conversation in Example 15. Asking ChatGPT to propose semantic traits that should be included in the list}
\label{example:16}

\noindent \framebox{\textbf{User}}\\
\noindent Are there any semantic traits that should be included that are currently absent from the list?\\

\noindent \framebox{\textbf{ChatGPT}}\\
\noindent To comprehensively define "radiative forcing" in the context of climate change, the list could be enhanced by including the following semantic traits that are currently absent or not explicitly mentioned:
 
\begin{itemize}
\item \textbf{Quantification of the effect of radiative forcing on climate systems:} While it's stated that radiative forcing is measured in watts per square meter, the direct impact on global temperature or climate change over time could be more explicitly addressed. This includes how specific quantities of radiative forcing correlate with temperature changes on Earth.
   
\item \textbf{Time scale of the impact of radiative forcing:} The list does not explicitly mention how the effects of radiative forcing might vary over different time scales, which is crucial for understanding its role in climate change dynamics.

\item \textbf{Comparative scale of different forcings:} An indication of the relative strength or importance of different types of radiative forcing (e.g., comparing the forcing from CO2 emissions to that from solar irradiance changes) would provide more context on the various sources and their impacts.

\item \textbf{Role of feedback mechanisms:} The interaction between radiative forcing and climate feedback mechanisms (like water vapor feedback, ice-albedo feedback) is crucial for a full understanding of the climate system's response to radiative forcing.

\item \textbf{Distinction between direct and indirect radiative forcing:} Clarifying the difference between direct radiative forcing (e.g., from CO2, solar irradiance) and indirect radiative forcing (e.g., changes in cloud cover, aerosol interactions) would provide a more nuanced understanding of the concept. 
\end{itemize}

\noindent Adding these traits would give a more complete and nuanced understanding of radiative forcing and its significance in the study of climate change.

\subsection*{Example 17: Continuation of the conversation in Example 16. Asking ChatGPT to write a new definition based on the list of semantic traits}
\label{example:17}

\noindent \framebox{\textbf{User}}\\
\noindent Write a new definition of "radiative forcing" for a terminological resource about climate change based on the list of semantic traits. Justify your definition and list the sources.\\

\noindent \framebox{\textbf{ChatGPT}}\\
\noindent Radiative forcing is a measure of the change in the energy balance of the Earth's climate system, quantified as the difference in watts per square meter (W/m\^2) between incoming solar radiation and outgoing infrared radiation at the top of the troposphere, particularly at the tropopause. It is a key factor in climate dynamics, determining the Earth's surface temperature by either warming or cooling the planet. Positive radiative forcing, often resulting from increased concentrations of greenhouse gases, enhances the greenhouse effect by trapping more infrared radiation emitted from the Earth, leading to a net increase in absorbed energy and a warming effect. Conversely, negative radiative forcing, which can be caused by factors like increased aerosol concentrations or changes in surface albedo, results in a net loss of energy and a cooling effect. The magnitude of radiative forcing is central to predicting the rate and extent of climate change, as it influences both the immediate energy balance and long-term climate feedback mechanisms.\\

\noindent \textbf{Justification:}
\begin{itemize}
\item \textbf{Energy balance focus:} The definition emphasizes the concept of energy balance, which is central to understanding radiative forcing and its climatic implications. 

\item \textbf{Quantification and location:} Specifying the measurement in watts per square meter at the top of the troposphere provides a clear, quantifiable description that aligns with scientific standards. 

\item \textbf{Cause and effect relationship:} It addresses the cause (greenhouse gases, aerosols, albedo changes) and effect (warming or cooling) directly, highlighting the role of radiative forcing in climate change. 

\item \textbf{Feedback mechanisms consideration:} Although not explicitly detailed in the traits list, the mention of climate feedback mechanisms acknowledges the complex interactions within the climate system that radiative forcing affects.\\ 
\end{itemize}
\newpage
\noindent \textbf{Sources:}
\begin{itemize}
\item EIA, IPCC: for the concept of radiative forcing occurring at the top of the troposphere and its quantification. 

\item UNFCCC, CARA, EO-NASA, IPCC, NAL: for the explanation of how radiative forcing alters the balance between incoming and outgoing radiation. 

\item Climate CoLab, GCMD-NASA, CCCR: for the broader context of radiative forcing's role in the climate system and its measurement units. 

\item OSU: for the distinction between positive and negative radiative forcing and its impact on global climate change.
\end{itemize}

\subsection*{Example 18: Asking ChatGPT to analyze some concordance lines}
\label{example:18}

\noindent \framebox{\textbf{User}}\\
\noindent Analyze the following concordance lines of the term "biological phosphorus removal" from a semantic point of view. What semantic traits of the term can you extract from them? Always mention the lines to justify your answer.
 
\begin{enumerate}
\item The parent sequencing batch reactor was operated over 100 d with synthetic wastewater (see Supporting Information for details) and achieved stable biological phosphorus removal efficiency (.99\%). 

\item By contrast, metagenomics of reactors with certain intentionally enhanced functions, for example, enhanced biological phosphorus removal reactor (EBPR), cellulose-degrading reactor, phenol decomposing reactor, sludge digester etc., makes more practical sense for most research institutions lack of such admirable resources, and thus is crucial for wide application of metagenomic techniques. 

\item Key words: Bioreactor, biofilm carriers, nitrogen removal, biological phosphorus removal, sewage, treatment reliable for organic carbon and nutrients removal without some of the problems of activated sludge processes[5]. 

\item Fig. 3: The average effluent soluble COD concentration from each reactor during experiment Phosphorus removal: Biological P-removal using enhanced biological phosphorus removal (EBPR) was carried out in this study. 

\item Phosphorus can also be removed through enhanced biological phosphorus removal (EBPR), however, the successful operation of EBPR depends on many process operational factors, especially variations in wastewater quality [14]. 

\item Biological phosphorus removal (BPR) [2], chemical precipitation, and adsorption by functionalized materials [3] are commonly used for phosphorus removal. 

\item In Concept 1, additional mixing energy of 5 MJ/kg Premoved originated from the biological phosphorus removal [7]. 

\item Another practical aspect worth highlighting in this section is the use of an enhanced biological phosphorus removal (EBPR) for phosphorus removal from wastewaters. 

\item " (12) "Degradation of diuron by Phanerochaete chrysosporium: role of ligninolytic enzymes and cytochrome P450" by J. da Silva Coelho-Moreira et al. (13) "Dynamics of intracellular polymers in enhanced biological phosphorus removal processes under different organic carbon concentrations" by L. 

\item Biological Treatment of Wastewater by Sequencing Batch Reactors such as biological phosphorus removal, aerobic oxidation of nitrogen (nitrification) and anoxic elimination of nitrate (denitrification). 

\item In the biological phosphorus removal, the phosphorus in the influent wastewater is incorporated into cell biomass, which subsequently is removed from the process as a result of sludge wasting. 

\item Our approach included the creation and use of domain specific HMMs to find members of the type I PKS domain in six published metagenomic data sets - Minnesota farm soil (MSF) [21], Sargasso Sea (SGS) [22], human gut (HGUT) [23], acid mine drainage (AMD) [24], enhanced biological phosphorus removal sludges (EBPRS) [25] and whale falls (bones from sunken whales) (WLF) [21]. 

\item In nearly all seven data sets the KS domain is found most frequently (with the exception of enhanced biological phosphorus removal sludge data sets) followed by the AT, PP or KR domains. 

\item A2O-activated sludge systems were composed of an anaerobic/anoxic/aerobic process with simultaneous nitrification-denitrification and biological phosphorus removal . 

\item Biological phosphorus removal in wastewater treatment systems using the enhanced biological phosphorus removal (EBPR) process is increasingly being used as an alternative to chemical precipitation processes [8]. 

\item Biological phosphorus removal in wastewater treatment systems using the enhanced biological phosphorus removal (EBPR) process is increasingly being used as an alternative to chemical precipitation processes [8]. 

\item Both plants are configured for biological phosphorus removal and have stable performance. 

\item Determination was made in the first stage that Lepothrix sppwas present at 30 °C, while 20° acetate was consumed in anaerobic condition and biological phosphorus removal by Rhodocyclacae-affiliating OUT 14 at 30 and 35° Celsisus acetate was completed and phosphorus removal did not occur as high [71]. 

\item WWTPs can be engineered to enable enhanced biological phosphorus removal (EBPR). 

\item Engineered systems performing enhanced biological phosphorus removal (EBPR), offer stable, lower complexity communities for studying phage-host interactions. 

\item One such system, wastewater treatment, relies on a process known as enhanced biological phosphorus removal (EBPR; [4]) to remove dissolved organic carbon and phosphorus. 

\item Gram-positive Tetrasphaera-related organisms (Actinobacteria) are putative polyphosphate-accumulating organisms that are abundant in many fullscale enhanced biological phosphorus removal plants [55] and Haliscomenobacter are one of the filamentous bacteria detected in wastewater treatment plants which are responsible for biomass bulking [56,57]. 

\item Nowadays, enhanced biological phosphorus removal (EBPR) has been broadly applied in wastewater containing phosphorus (P) treatment because of its lower cost and more environmental friendly technology compared to chemical precipitation or adsorption [1]. 

\item The P content of excess sludge in this study, namely, anoxic sludge, was 12.3\% of dry weight significantly higher than that in traditional biological phosphorus removal process (3\%) [18], which agrees well with the results of chemical analysis, further demonstrating the existence of phosphorus removal microorganisms in this two-sludge process studied here. 3.2. 

\item Wilmes et al. conducted a series of metaproteomic studies on the molecular mechanisms of enhanced biological phosphorus removal (EBPR) [4,21,22].\\
\end{enumerate}

\noindent \framebox{\textbf{ChatGPT}}\\
\noindent From the provided concordance lines, we can extract several semantic traits of "biological phosphorus removal" (BPR):

\begin{enumerate}
\item \textbf{Process Efficiency and Stability}: Line 1 mentions a reactor achieving "stable biological phosphorus removal efficiency," indicating the process's reliability and effectiveness. 

\item \textbf{Comparative Advantage}: Lines 2 and 23 highlight the practicality and environmental preference of enhanced biological phosphorus removal (EBPR) over other methods, indicating a comparative advantage in terms of resource usage and environmental impact. 

\item \textbf{Application in Wastewater Treatment}: Many lines (e.g., 3, 10, 11, 15, 17, 19, 21) associate biological phosphorus removal with wastewater treatment, revealing its primary application in removing phosphorus from wastewater to prevent pollution. 

\item \textbf{Integration with Other Processes}: Lines like 3, 14, and 20 mention the integration of biological phosphorus removal with other processes like nitrogen removal and nitrification-denitrification, suggesting its compatibility and combinatory use in comprehensive waste treatment systems. 

\item \textbf{Dependency on Operational Factors}: Line 5 and 18 discuss how the success of EBPR depends on operational factors and conditions, such as wastewater quality and temperature, indicating a sensitivity to environmental and operational variables. 

\item \textbf{Microbial Activity}: Lines like 11, 22, and 24 refer to the role of microorganisms in phosphorus removal, highlighting the biological aspect of the process where specific bacteria or microbial communities are essential for phosphorus uptake and accumulation. 

\item \textbf{Technological and Research Interest}: Lines 2, 9, 12, and 25 show the interest in researching and enhancing the understanding of the biological mechanisms and efficiency of phosphorus removal, indicating ongoing scientific and technological development in this field. 

\item \textbf{Economic and Environmental Benefits}: Line 23 specifically mentions the lower cost and environmental friendliness of EBPR, portraying it as a sustainable and cost-effective solution for phosphorus removal in wastewater treatment. 
\end{enumerate}

\noindent These lines collectively suggest that biological phosphorus removal, particularly in its enhanced form (EBPR), is a crucial, efficient, and environmentally preferred method for treating wastewater, with ongoing research and development focused on optimizing its application and understanding the underlying biological processes.

\subsection*{Example 19: Asking ChatGPT to analyze some concordance lines where the target term is the object of the sentence}
\label{example:19}

\noindent \framebox{\textbf{User}}\\
\noindent Analyze the following concordance lines of the term "hash" from a semantic point of view. What semantic traits of the term can you extract from them? Always mention the lines to justify your answer. 

\begin{enumerate}
\item If one person in your pool succeeds at completing the hash, you split the value of the Bitcoin. 

\item Changing the hash stored in block 3 changes the overall data stored inside it. 

\item After a unique hash is found (below the target), miners notify the network so that others update their information in the blockchain as well. 

\item Signing a hash allows the total operation to be much faster, since signing long messages can be very slow. 

\item If we wanted to make sure that we're looking at the same data that was originally hashed, it would be easy as long as we were using the same program to create our hashes. 

\item "OP\_EQUAL" (not shown) checks the two values at the top of the stack; in this case, it checks whether the pubkey hash generated from the full public key Bob provided equals the pubkey hash Alice provided when she created transaction \#1. 

\item until we find a hash that starts with a single zero. 

\item Whenever you make a payment in cryptocurrency, you receive a hash displayed in your wallet. 

\item His full (unhashed) public key, so the pubkey script can check that it hashes to the same value as the pubkey hash provided by Alice. 

\item The motivation for using the public key hash was to shorten the size of the address. 

\item But now, the validator script that I created requires as one of it's parameters a Plutus associating public key hashes to integers. 

\item This means the hash is often denoted as a combination of numbers (0 to 9) and letters (a to f). 

\item Nodes on the blockchain network can independently verify the hash of each block in the chain, ensuring that the data has not been altered. 

\item I gave "Bitcoin Seed" as key and generated a hash, but it is not matching the output showed on Ian Coleman's bip39 website. 

\item Calculate script hash of the contract (20 bytes, make once SHA256 and RIPEMD160 of the script). 

\item N.B. Smart contracts can verify data authenticity and integrity using its hash, avoiding the need for direct on-chain storage, similar to how NFT metadata is stored on IPFS while only its hash is sent to the blockchain. 

\item Where can you find public key hashes ? 

\item In the table above, the SHA-256 hash function is used to generate the hashes of three different inputs. 

\item Mining entails a plethora of hashing attempts until a valid hash is created . 

\item Beginning with the public key, this string of values is first ran through the SHA-256 hashing algorithm to produce a hash, and then that hash is computed using RIPEMD160 to produce a bitcoin address. 

\item However, if you input the same data again, it will always produce the same hash . 

\item Since the hash is generated before the signature is computed, we need to exclude the signature block of a certificate when computing the hash. 

\item Whenever you make a payment in cryptocurrency, you receive a hash displayed in your wallet. 

\item This means the hash is often denoted as a combination of numbers (0 to 9) and letters (a to f). 

\item Changing the hash stored in block 3 changes the overall data stored inside it.\\
\end{enumerate}

\noindent \framebox{\textbf{ChatGPT}}\\
\noindent Analyzing these lines, several semantic traits of the term "hash" emerge, which relate to its role in technology and cryptography:
 
\begin{enumerate}
\item \textbf{Cryptocurrency and Blockchain Context:} Lines 1, 3, 8, 9, 17, 19, and 23 associate "hash" with the process of cryptocurrency transactions and blockchain technology. The hash functions as a unique identifier or verification tool in these systems.

\item \textbf{Data Integrity and Verification:} Lines 2, 5, 13, 14, 15, and 16 suggest that a hash is used to ensure data integrity and verification. For example, line 13 mentions verifying the hash of each block to ensure data has not been altered, highlighting its role in maintaining the accuracy and consistency of data.

\item \textbf{Unique and Consistent Output:} Lines 21 and 22 imply that a hash function produces a unique and consistent output for the same input data, which is fundamental in cryptographic processes and data verification.

\item \textbf{Speed and Efficiency:} Line 4 mentions that signing a hash can be much faster than signing long messages, indicating the efficiency and speed that hashing brings to cryptographic operations.

\item \textbf{Compact Representation:} Lines 6, 10, and 20 discuss how hashes can shorten the size of data or addresses, suggesting that a hash serves as a compact representation of larger sets of data.

\item \textbf{Mining Process:} Lines 7 and 19 refer to the process of mining, where a specific hash value is sought, demonstrating the hash's role in the computational effort to validate blockchain transactions.

\item \textbf{Digital Signatures and Authentication:} Lines 6, 9, and 22 highlight the use of hashes in creating and verifying digital signatures, which are crucial for authentication in cryptographic protocols.

\item \textbf{Alphanumeric Representation:} Lines 12 and 24 indicate that hashes are often represented as a combination of numbers and letters, showing their format as alphanumeric strings.

\item \textbf{Technical Implementation:} Lines 18 and 20 mention specific hashing algorithms like SHA-256 and RIPEMD160, pointing to the technical aspects of how hashes are generated and their role in cryptographic functions.

These lines collectively portray "hash" as a fundamental concept in cryptography and digital transactions, emphasizing its importance in ensuring data integrity, facilitating efficient verification, and enabling secure and authenticated communication.
\end{enumerate}

\subsection*{Example 20: Asking ChatGPT to analyze some concordance lines where the target term holds a causal relation with other terms}
\label{example:20}

\noindent \framebox{\textbf{User}}\\
\noindent Analyze the following concordance lines of the term "erosion" from a semantic point of view. What semantic traits of the term can you extract from them? Always mention the lines to justify your answer.

\begin{enumerate}
\item Coastal erosion in Europe causes significant economic loss, ecological damage and societal problems. 

\item Thus, dune erosion generally leads to a permanent loss of sand which can only be compensated by artificial nourishment (dune restoration). 

\item Without tectonic uplift, erosion could eventually produce a flat erosion surface, almost a plain (a peneplain). 

\item Many mountains are undoubtedly formed by erosion of earlier plateaux, and even the Andes, a stronghold of plate tectonic explanations, are dominated by the altiplano (high plains ). 

\item Excessive levels of sedimentation are caused by activities such as soil erosion from agriculture, dredging, and drilling for oil and gas. 

\item Abyssal plains are covered by fine-grained sediment derived from continental erosion . 

\item Both tectonic denudation and subaerial erosion may lead to the unroofing of deeply-buried rock, which will expand once the confining pressure is removed. 

\item The lateral thermal erosion produces a thermo-erosional niche . 

\item Examine how the processes of erosion and deposition cause barrier islands to migrate. 

\item The net rate is a calculation affected by sand sources and sinks as well as the change in shoreline orientation resulting from erosion . 

\item The most critical problem facing the grunion resource is the loss of spawning habitat caused by beach erosion . 

\item Downdrift erosion may well lead to pressure for further defence works. 

\item The sediment generated by erosion may accumulate in the immediate vicinity or it may be transported elsewhere. 

\item Continuing shoreface erosion may ultimately lead to an increased wave attack intensifying the transport capacity and hence intensified erosion (negative feed-back system). 

\item Coastal erosion causes significant economic loss, ecological damage and societal problems. 

\item In a few cases, best illustrated by the example of Cap Ferret, public authorities are even held responsible for damage induced by coastal erosion because they granted construction permits in areas at risk. 

\item Sediments result from soil erosion in water catchments and are transported to rivers by water run-off. 

\item Erosion problems are often located at urbanized areas because erosion causes the largest damage in these areas. 

\item Section C is characterized by coastal erosion which caused settlement dislocation. 

\item Fig. 12 shows how coastal erosion leads to building damage and loss. 

\item Erosion causes sedimentation in lakes, reservoirs, and streams, which eventually require dredging. 

\item This erosion will cause higher waves at the toe of the new seawall, with consequently higher overtopping discharges. 

\item This assumption is likely reasonable when the local slope change induced by erosion/deposition of sediments is small compared to the mean beachface slope. 

\item Erosion generally leads to either great monetary losses due to storm damage or even larger expenditures for shore protection to prevent the loss. 

\item This book examines the problems caused by soil erosion from farms, which is the main cause of nonpoint source pollution.\\
\end{enumerate}

\noindent \framebox{\textbf{ChatGPT}}

\noindent Analyzing these concordance lines of the term "erosion," we can extract several semantic traits: 

\begin{enumerate}
\item \textbf{Destructive Nature}: Erosion is frequently associated with damage and loss. Lines 1, 15, 19, and 20 mention economic loss, ecological damage, societal problems, and settlement dislocation, emphasizing the destructive impact of erosion on both the environment and human infrastructure.

\item \textbf{Geological Process}: It involves the wearing away of earth or rock, particularly in the context of coastal, soil, and tectonic activities. Lines 3, 4, and 6 speak of geological processes like the formation of flat surfaces, mountain formation, and sediment coverage on abyssal plains.

\item \textbf{Human-Induced Factors}: Erosion is often exacerbated by human activities such as agriculture, construction, and industrial operations. Lines 5, 16, and 25 mention soil erosion from agriculture, construction permits in risk areas, and nonpoint source pollution from farms.

\item \textbf{Need for Intervention}: There is a recurring theme of the need for human intervention to manage or mitigate erosion effects, such as artificial nourishment, dune restoration, and shore protection mentioned in lines 2, 10, and 24.

\item \textbf{Resultant Features and Processes}: Erosion leads to specific geological and environmental features or processes, such as the creation of peneplains, altiplanos, and sedimentation in water bodies, as well as the unroofing of rocks (lines 3, 4, 7, and 21).

\item \textbf{Transportation of Materials}: Erosion involves the movement of soil or rock materials, either leading to accumulation in new areas or the loss from the original site, as discussed in lines 13 and 17.

\item \textbf{Feedback Mechanisms}: It can be part of a feedback system where erosion processes intensify other processes, such as increasing wave attack or causing higher waves, as mentioned in lines 14 and 22.

\noindent These traits collectively describe erosion as a significant, dynamic process that impacts landscapes, ecosystems, and human activities, often requiring management or mitigation to prevent or reduce its detrimental effects.
\end{enumerate}

\subsection*{Example 21: Asking ChatGPT to analyze some concordances lines and then writing a definition based on them}
\label{example:21}

\noindent \framebox{\textbf{User}}

\noindent Analyze the following concordance lines of the term "false seedbed technique" from a semantic point of view. What semantic traits of the term can you extract from them? Always mention the lines to justify your answer. 

\begin{enumerate}
\item Authors compared a conventional weed control system which relied on the use of chemical and mechanical methods with an alternative system based on the use of false seedbed technique and precision weeder. 

\item PRACTICE ABSTRACT NO.007 Reducing weed seed pressure with the false seedbed technique Problem Applicability box Annual crops are especially sensitive to weed pressure during early growth. 

\item Reducing weed seed pressure with the false seedbed technique . 

\item The importance of shallow tillage as a weed control method in the false seedbed technique has also been highlighted. 

\item In the false seedbed technique seedling destruction usually occurs by harrowing or similar mechanical tools whereas in the case of the stale seedbed technique it occurs by chemical herbicides or by thermal methods (flame weeding or soil steaming), to avoid any further soil disturbance. 

\item The combined effects of false seedbed technique, post-emergence chemical control and cultivar on weed management and yield of barley in Greece. 

\item Weed management in peanut using false seedbed techniques . 

\item Seedbed preparation is the final secondary tillage operation except when used in the stale or false seedbed technique (Leblanc and Cloutier, 1996). 

\item Predicting the main flush of weeds in the field could maximize the efficacy of false seedbed technique as weed management practice. 

\item The review described the influences of timing and type of tillage on the promotion of weed emergence and on the devitalization of emerged weeds, which are the basis of the false seedbed technique . 

\item Weed control strategies involved the use of biodegradable mulch together with mechanical weeding, or false seedbed technique and mechanical weeding. 

\item The false seedbed technique has several limitations. 

\item The main objective of this study was to gain insight into how soil tillage (here, the false seedbed technique ) and cover crops impact spontaneous plant communities in fallows preceding banana cropping. 

\item The false seedbed technique has not been well documented; however, this practice is widespread on organic farms (Mohler, 2001) and a reduction in weed density of 63–85\% has been observed in some situations (Gunsolus, 1990; Leblanc and Cloutier, 1996). 

\item Reducing weed seed pressure with the false seedbed technique . 

\item Despite its limitations, the false seedbed technique has broad applicability. 

\item When the false seedbed technique was used (M2), the 5-cm upper soil layer was fragmented three times before sowing at 5 cm depth with a seedbed cultivator and a power harrow. 

\item Our aim was to gain insight into how soil tillage (here, the false seedbed technique ) and cover crops impact the spontaneous vegetation in fallows preceding banana crops. 

\item False seedbed technique aims to reduce weed seed bank by exploiting seed germination biology. 

\item The system utilized for physical weed control was based first on a false seedbed technique followed by pre-sowing weed removal, performed with a special 2 m wide 6-row spring-tine harrow. 

\item The false seedbed technique reduces the weed seed bank in the topsoil and, as a result, significantly reduces competition of annual weeds in the succeeding crop. 

\item Moreover, the false seedbed technique may have failed to destroy rhizomes present in the superficial layers (Lamour and Lotz 2007; Ringselle et al. 2019). 

\item The importance of shallow tillage as a weed control method in the false seedbed technique has been highlighted. 

\item False seedbed technique is a method providing weed seed bank depletion. 

\item Predicting weed emergence can answer the question of how much time weed control and crop sowing should be delayed in a specific agricultural area where false seedbed technique is about to be applied.
\\

\end{enumerate}

\noindent \framebox{\textbf{ChatGPT}}

\noindent Analyzing the concordance lines of the term "false seedbed technique," several semantic traits emerge: 

\begin{enumerate}
\item \textbf{Alternative Weed Control Method}: Lines 1 and 14 indicate that the false seedbed technique is an alternative to conventional weed control methods, suggesting its use as a sustainable or less chemically dependent approach. 

\item \textbf{Mechanical Weed Suppression}: Lines 5, 7, and 20 describe the technique as primarily mechanical, involving harrowing or similar tools for seedling destruction, contrasting with chemical or thermal methods in other techniques. 

\item \textbf{Weed Seed Bank Reduction}: Lines 2, 19, 21, and 24 mention the goal of reducing weed seed pressure or depleting the weed seed bank, indicating the technique's focus on preventing weed growth by targeting the seeds before they can establish. 

\item \textbf{Soil Tillage Component}: Lines 4, 8, 10, 17, and 23 highlight shallow tillage as a crucial element of the technique, emphasizing soil manipulation to encourage weed seed germination before destroying them. 

\item \textbf{Timing and Planning}: Lines 9, 10, and 25 suggest that the technique's success depends on timing, particularly in relation to weed emergence and tillage operations, to maximize weed control efficacy. 

\item \textbf{Applicability and Limitations}: Lines 12, 14, 16, and 22 discuss the technique's broad applicability, especially in organic farming, while also acknowledging its limitations, such as potential failure to destroy certain weed parts like rhizomes. 

\item \textbf{Integrated Weed Management}: Lines 6, 11, and 20 imply that the false seedbed technique is often part of a broader weed management strategy, integrating mechanical, cultural, and sometimes chemical methods. 
\end{enumerate}

\noindent These traits collectively characterize the false seedbed technique as a proactive, mechanical, and soil-focused method of weed control, emphasizing the strategic timing and integration within broader weed management practices.\\

\noindent \framebox{\textbf{User}}\\
\noindent Write a definition of "false seedbed technique" ready to be inserted in a terminological resource based the concordance lines. Include a justification of the definition mentioning the lines on which you have based your definition.\\

\noindent \framebox{\textbf{ChatGPT}}
\\
\noindent \textbf{Definition:} 

\noindent The false seedbed technique is a weed management method that involves the shallow tillage of soil to stimulate the germination of weed seeds, followed by their destruction through mechanical means such as harrowing or using a spring-tine harrow. This technique aims to reduce the weed seed bank in the topsoil, thereby decreasing the competition from annual weeds in subsequent crops.
\\\\
\noindent \textbf{Justification:}

\noindent This definition synthesizes information from several key points in the concordance lines:

\begin{itemize}

\item \textbf{Shallow Tillage and Mechanical Destruction}: Lines 4, 5, and 23 emphasize the importance of shallow tillage as a weed control method within the false seedbed technique and the use of mechanical tools for seedling destruction. 

\item \textbf{Weed Seed Bank Reduction}: Lines 2, 19, 21, and 24 specifically mention the objective of reducing weed seed pressure or depleting the weed seed bank, highlighting the technique's preventive approach to weed control. 

\item \textbf{Weed Management Strategy}: Lines 1, 14, and 16 illustrate the technique's role as an alternative weed control method, often utilized in sustainable or organic farming contexts due to its mechanical nature and minimal soil disturbance. 
\end{itemize}

\noindent This comprehensive view from the concordance lines ensures the definition captures the essence of the false seedbed technique as a proactive, mechanical weed management practice.

\subsection*{Example 22: Asking ChatGPT to evaluate a definition in search of different types of errors}
\label{example:22}

\noindent \framebox{\textbf{User}}\\
\noindent Evaluate this definition in search of the following issues: 1) spelling mistakes, 2) poor stylistic choices, 3) bias, 4) factual errors. Finally, rewrite the definition taking into account your own suggestions.
\\\\
\noindent nuclear energy: energy derived from the nuckleus of an atom commonly harnessed through two methods: fission and fusion. Fission involves the splitting of heavy atomic nuclei into smaller particles, releasing a significant amount of energy. This method is predominantly used in nuclear power plants to generate electricity. Fusion, on the other hand, involves the combination of light atomic nuclei to form a heavier nucleus, releasing energy in the process. Fusion is the power source of the sun and stars but has not yet been commercially realized on Earth due to the immense conditions required. Nuclear energy is a safe source of low-carbon power and offers a solution to energy demands with reduced greenhouse gas emissions.\\

\noindent \framebox{\textbf{ChatGPT}}
\\
1) \textbf{Spelling Mistakes:} 
\begin{itemize}
    \item "nuckleus" should be "nucleus". 
\end{itemize}
 
\noindent 2) \textbf{Poor Stylistic Choices:} 
\begin{itemize}
   \item The definition starts directly with "energy derived from the nucleus of an atom" which could be prefaced for clarity. 

   \item The phrase "which a tremendous amount of energy" is awkward and seems incomplete. 
\end{itemize}
 
\noindent 3) \textbf{Bias:}
\begin{itemize}
   \item The statement "Nuclear energy is a safe source of low-carbon power" presents a positive view of nuclear energy without acknowledging the associated risks and controversies.
\end{itemize}

\noindent 4) \textbf{Factual Errors:} 
\begin{itemize}
    \item Fission involves the splitting of *heavy* atomic nuclei, not light atomic nuclei. 
    \item Fusion involves the combination of *light* atomic nuclei, not heavy atomic nuclei, to form a heavier nucleus.
\end{itemize}

\noindent \textbf{Rewritten Definition:} 

\noindent Nuclear energy refers to the power generated from the nucleus of an atom, primarily through two processes: fission and fusion. Fission is the process of splitting heavy atomic nuclei into smaller particles, releasing a significant amount of energy. This method is predominantly used in nuclear power plants to generate electricity. Fusion, conversely, is the merging of light atomic nuclei to create a heavier nucleus, a process that liberates energy and powers the sun and other stars. Although fusion promises a substantial energy yield, it has not been commercially harnessed on Earth due to the extreme conditions required for its initiation. While nuclear energy contributes to low-carbon power generation and can meet substantial energy demands, its safety and environmental impact are subjects of ongoing debate and research.

\subsection*{Example 23: Asking ChatGPT to guess the defined term and improve the definition}
\label{example:23}

\noindent \framebox{\textbf{User}}\\
What is the term defined by this definition?

\noindent Any process, activity or mechanism that removes carbon dioxide from the atmosphere.\\

\noindent \framebox{\textbf{ChatGPT}}\\
The term defined by this definition is "carbon sequestration." It refers to the process, activity, or mechanism that removes carbon dioxide (CO2) from the atmosphere and stores it in a carbon sink, which can be either natural, like forests and oceans, or artificial, like carbon capture and storage technology.\\

\noindent \framebox{\textbf{User}}\\
Why is it not a definition of "carbon dioxide sink"? Rewrite the definition so that it defines "carbon dioxide sink".\\

\noindent \framebox{\textbf{ChatGPT}}\\
The original definition refers more to the process of removing and storing carbon dioxide, rather than the place or object where the carbon dioxide is stored. A "carbon dioxide sink" refers to a reservoir, either natural or artificial, that accumulates and stores some carbon-containing chemical compound for an indefinite period. 

\noindent To redefine the term to specifically refer to "carbon dioxide sink," the definition would be: 

\noindent "A carbon dioxide sink is a natural or artificial reservoir that absorbs and stores carbon dioxide from the atmosphere, thereby reducing the concentration of CO2 and contributing to climate change mitigation."\\

\end{document}